\documentclass[11pt,a4paper]{article}
\usepackage[hyperref]{emnlp-ijcnlp-2019}
\usepackage{times}
\usepackage{latexsym}
\usepackage{amssymb}

\usepackage{graphicx}
\usepackage{url}

\aclfinalcopy 

\title{Decoupled Box Proposal and Featurization with \\Ultrafine-Grained Semantic Labels Improve \\Image Captioning and Visual Question Answering}

\author{Soravit Changpinyo~~~~~~~Bo Pang~~~~~~~Piyush Sharma~~~~~~~Radu Soricut\\
  Google AI \\
  Venice, CA 90291 \\
  {\tt \{schangpi,bopang,piyushsharma,rsoricut\}@google.com} \\
  }

\date{}

\begin{document}
\maketitle

\begin{abstract}
 Object detection plays an important role in current solutions to vision and language tasks like image captioning and visual question answering.  However, popular models like Faster R-CNN rely on a costly process of annotating ground-truths for {\em both} the bounding boxes {\em and} their corresponding semantic labels, making it less amenable as a primitive task for transfer learning. In this paper, we examine the effect of decoupling box proposal and featurization for down-stream tasks. The key insight is that this allows us to leverage a large amount of labeled annotations that were previously unavailable for standard object detection benchmarks. Empirically, we demonstrate that this leads to effective transfer learning and improved image captioning and visual question answering models, as measured on publicly-available benchmarks.
 \end{abstract}

\section{Introduction}

Object detection has been employed extensively as a primitive task for vision and language tasks such as image captioning and visual question answering (VQA); see \cite{anderson18bottomup} and the work that follows it.
One motivation is that the ability to recognize salient regions and objects may be too difficult to learn from weakly-supervised \emph{top-down} signals, in the form of captions and question-answer pairs.
Indeed, \emph{bottom-up} signals provided by object detection often correspond to semantic units of language such as words or phrases, making them suitable for text generation and image-text alignment.

However, object detection itself can be broken down into multiple subtasks \cite{liu18deep}.
A family of ``two-stage" object detectors first proposes category-agnostic bounding box candidates and then featurizes and classifies the cropped regions into one of the available semantic labels.
Even ``one-stage" object detection approaches, where these boxes become category-specific, can be formulated in a \emph{bottom-up} manner as detecting and grouping extreme and center points \cite{zhou19bottom}.
Can we take advantage of this observation to learn to transfer more effectively?
In this work, we take a step in this direction by examining the effect of decoupling box proposal and featurization \emph{on downstream vision and language tasks}.
In particular, we consider a two-stage object detector and set a goal of pushing the ``featurization" aspect of the task further than before.

\begin{figure}[t]
\begin{center}
 \includegraphics[width=1\linewidth]{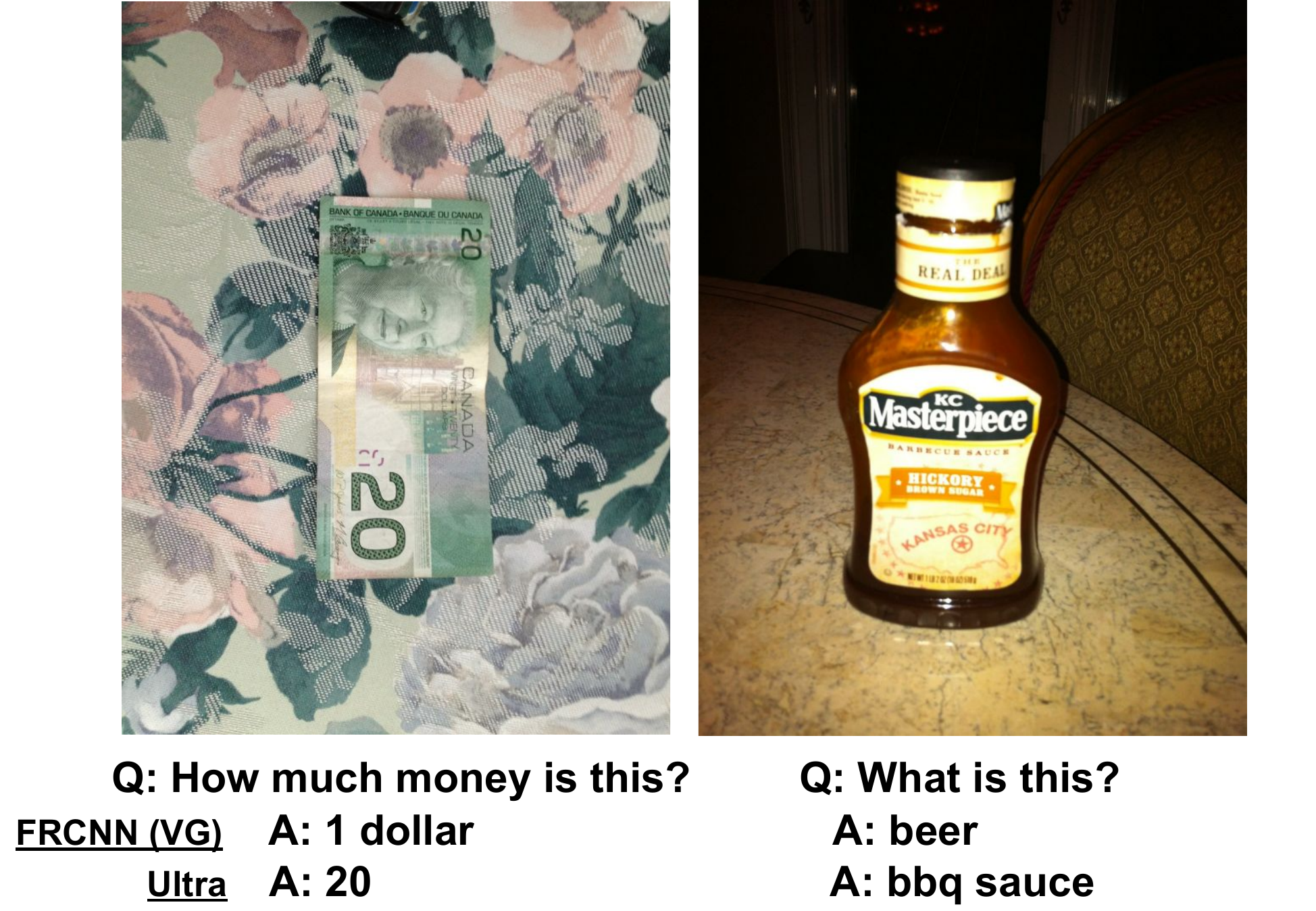}
\end{center}
 \vspace{-6pt}
 \caption{Ultrafine-grained semantic labels (at ``instance level'') provide transfer learning power to downstream tasks like visual question answering.}
\label{fig:examples}
\vspace{-12pt}
\end{figure}

Our choice to break free from ``featurization by object detection models" has at least two advantages.
First, there is a larger amount of labeled data that can be leveraged to train a better featurization module, even if such data do not support learning box proposals.
To put it another way, the quality of features directly provided by object detectors is limited by the fact that annotating ground-truths for \emph{both} the bounding boxes {\em and} their corresponding semantic labels is costly and scales poorly.
By separating them, we reintroduce the freedom to annotate for object-agnostic box segmentation, without the burden of baking in annotation decisions related to the granularity level of the semantic labels (i.e., do we use as semantic labels ``money'', ``euro'', or ``20 euro''?).
As illustrated in Figure~\ref{fig:examples}, the granularity level of the semantic labels plays a crucial role for downstream tasks such as VQA.

Second, this approach is better suited to downstream tasks whose domains are different from the one the object detector is trained on.
In other words, it allows us to benefit from transfer learning, which is a great advantage given the relatively modest amount of available supervised data for these downstream vision and language tasks.

We empirically demonstrate the above-mentioned advantages through a focused study of the effect of improved featurization on image captioning and VQA in transfer learning settings.
In particular, we (i) leverage ultra-fine-grained semantic labels (e.g., ``golden gate bridge'' vs. ``bridge'') for featurization \cite{juan19graphrise};
and, (ii) focus on scenarios in which object detection modules trained on Visual Genome (VG) \cite{krishnavisualgenome} are applied to out-of-domain images:
image captioning on the Conceptual Captions dataset \cite{sharma2018conceptual}, and VQA on the VizWiz dataset \cite{gurari18vizwiz}.
Our results indicate that there are ways to incorporate low-level pre-training tasks that benefit vision and language models via higher-quality bottom-up signals.

\section{Related Work}

Attention-based deep models are popular in image captioning and VQA.
Early work used fixed partitions of images as candidate regions \cite{xu15show}.
However, variable sized regions that are better correlated with object boundaries have gained momentum \cite{fu17aligning,pedersoli17areas,anderson18bottomup}.
Indeed, \newcite{anderson18bottomup} established new state-of-the-art performance over
both image captioning and VQA tasks on the MSCOCO and VQA2 benchmarks using a Faster R-CNN detector trained on Visual Genome.
As both Visual Genome and VQA2 were built on images from MSCOCO, the object detector was applied largely to in-domain images.
In contrast, our work focuses on more realistic settings in which domains of different tasks may not be perfectly aligned \cite{chen18domain}.

We leverage image representations extracted from
a network pre-trained over large amounts of labeled data.
Prior work demonstrated the power of pre-training with image classification at scale \cite{sun17revisiting,mahajan18exploring,wu19tencent}.
However, we consider downstream vision \emph{and language} tasks (image captioning and visual question answering), in contrast to less complex
vision-only tasks explored in such work: object detection and in some cases semantic segmentation and human pose estimation.
Furthermore, our transfer learning technique is based on decoupled region proposal and ultra-finegrained featurization, not fine-tuning the pre-trained network.

Another set of closely related work utilized additional data for scaling up either vision tasks \cite{hoffman16large,tang17visual,redmon17yolo} or vision and language tasks \cite{venugopalan17captioning,lu18neural,noh19transfer}.
For instance, YOLO9000 \cite{redmon17yolo} built a ``WordTree" hierarchy based on the WordNet synsets \cite{miller90wordnet}, mapped categories in both COCO object detection and ImageNet classification datasets into the hierarchy, and proposed a joint detection and classification training framework.
Our approach to transfer learning with ultrafine-grained featurization can similarly address the long-tail nature of target vocabulary (see Figure~\ref{fig:cap_examples}) while being simpler (e.g., not require carefully merging different sets of vocabulary as in YOLO9000).
The number of classes we consider is also several orders of magnitude larger.

Incorporating object detection signals in downstream tasks appropriately is non-trivial and an active subject for research \cite{santoro17relational,zhang18counter}.
In this work, we ask the orthogonal question of whether it is necessary to accept the object detector's output as-is.

\section{Features and Experimental Setup}
\label{s:setup}
Our starting point is a two-stage object detector, which consists of two core modules.
One is responsible for category-agnostic box proposal, and the other for featurizing each cropped region for semantic label prediction.
In this paper, we select Faster R-CNN \cite{ren15frcnn}, a widely-used object detector in image captioning and VQA.

\paragraph{Faster R-CNN Model}
We reimplement the Faster R-CNN model, training it to predict both 1,600 object and 400 attribute labels in Visual Genome \cite{krishnavisualgenome},
following the standard setting from \newcite{anderson18bottomup}.
ResNet-101 \cite{resnet2016} pre-trained on ImageNet \cite{imagenet15} is used as the core featurization network\footnote{See further details in the supplementary material.}.
We achieve a mAP@50 of 10.96 for object detection and 1.5 for attribute detection.
Given an image, Faster R-CNN proposes $\mathsf{K}$ bounding box regions, each of which comes with a $\mathsf{D}$-dimensional feature vector as well as object/attribute class predictions (along with their scores).
$\mathsf{K}$ is set to 100 and $\mathsf{D}$ to 2048 in our experiments.
Using output features on the task of VQA and our model described in Section~\ref{sec:vqa}, we obtain an accuracy of 66.9\% on the validation set of the VQA2 dataset \cite{goyal17vqa2}.
For comparison, this number already surpasses all validation accuracy numbers in Table 2 for a strong model by \newcite{peng18dynamic}, suggesting that our Faster R-CNN features are of high-quality.

\paragraph{Decoupled Box Proposal and Featurization with Ultra-finegrained Semantic Labels}
In standard use of object detectors following \newcite{anderson18bottomup}, downstream tasks receive ``knowledge'' merely about a few thousand classes and four hundred attributes.
Here, we exploit the fact that box proposal and featurization can be decoupled, and work on improving the object representation (featurization).

More concretly, we conduct a study toward understanding the utility of improved featurization on downstream tasks.
To this end, we exploit a graph-based, semi-supervised representation learning approach called Graph-Regularized Image Semantic Embedding (Graph-RISE) \cite{juan19graphrise}.
Specifically, Graph-RISE is based on ResNet-101 where the 10x10x2K feature map is first average pooled to 4x4x2K,
and then flattened and projected to a 64-dimensional embedding before the softmax layer.
Learned from $\mathcal{O}$(260M) web images and $\mathcal{O}$(40M) (noisy) semantic labels,
these compact 64-dimensional feature vectors are trained to capture a whole spectrum of semantic similarity,
ranging from coarse-grained / category-level (e.g., ``bridge''), fine-grained level (e.g., ``steel red bridge''),
to ultrafine-grained / instance-level (e.g., ``golden gate bridge'').

\paragraph{Our Objective}
The main goal is to compare two approaches in using bottom-up signals:
1) FRCNN: use the default visual features from the Faster R-CNN detector;
2) Ultra: use bounding boxes from the Faster R-CNN detector, then featurize them using the much more compact representation from Graph-RISE that potentially reflects differentiation of ultrafine-grained semantic labels.
Next, we evaluate this setup on downstream tasks for image captioning and visual question answering.

\section{Image Captioning}
\label{sec:ic}

\paragraph{Dataset}
We use the Conceptual Captions (CC) dataset~\cite{sharma2018conceptual}, consisting of 3.3 million training and 15,000 validation images/caption pairs.
Another 12,000 image/caption pairs comprise the hidden test set.
Official scores on the test set are obtained by submitting models to the CC Challenge server\footnote{ai.google.com/research/ConceptualCaptions}.
Unlike other image captioning datasets, images from CC are pulled from across the web and thus exhibit a wide variety of both images and image-caption styles.
Most notably, the domain of images can be very different from Visual Genome, unlike in popular benchmarks such as MSCOCO \cite{coco}.

\paragraph{Model}
We adopt the encoder-decoder model from \cite{sharma2018conceptual}, whose basic building block is a Transformer Network~\cite{vaswani2017attention}.
To convert multi-modal inputs to a sequence of encoder feature vectors,  we use up to three types of image features:
\begin{description}
\itemsep0em
\item[G]: Global features by Graph-RISE, a dense 64D vector extracted from the whole image; 
\item[B]: Box-region features by Faster R-CNN (FRCNN, sparse 2048D), or Graph-RISE (Ultra, dense 64D), extracted from each cropped image region resized to 224x224 (cf. Sec.~\ref{s:setup});
\item[L]: Label embeddings, obtained by embedding predicted object semantic labels from Google Cloud Vision APIs\footnote{cloud.google.com/vision} into a 512D feature vector.
These semantic labels are then mapped to embeddings using an embedding layer pre-trained to predict label co-occurrences in web documents using a word2vec model~\cite{mikolov-et-al:2013a}.
\end{description}

For both {\bf B} and {\bf L}, we select the inputs with highest scores and order the sequence inputs based on such scores from high to low.
Additionally for {\bf B}, we remove box regions whose scores are lower than 0.001.
We use beam search with width 5 for the decoder in all of our experiments\footnote{See further details in the supplementary material.}.

\paragraph{Metrics}
We 
adopt the standard automatic metrics for image captioning:
CIDEr~\citep{cider}, ROUGE-L~\cite{lin-och:2004}, and SPICE~\cite{spice}, as implemented in the COCO-caption evaluation toolkit\footnote{https://github.com/tylin/coco-caption.}.

\begin{figure*}[t]
\begin{center}
 \includegraphics[width=1.0\linewidth]{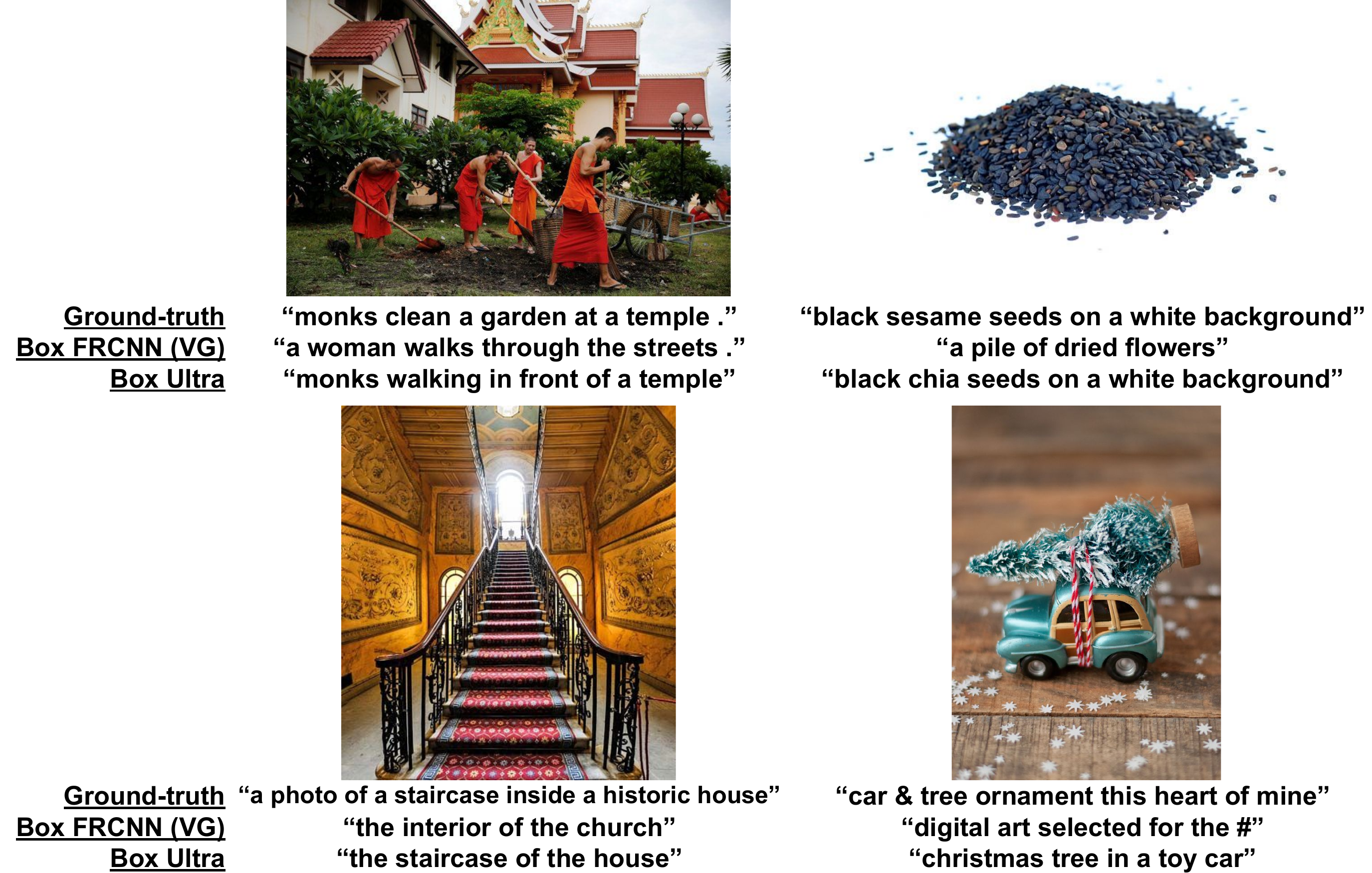}
\end{center}
 \vspace{-6pt}
 \caption{Qualitative results from our image captioning models using {\bf B}-FRCNN vs. {\bf B}-Ultra (see text for details), along with ground-truth captions.
 Ultra is more capable than FRCNN of dealing with images with unfamiliar objects,
 those that do not perfectly fall into the domain where the Faster R-CNN object detector is trained on.}
\label{fig:cap_examples}
\vspace{-12pt}
\end{figure*}

\begin{table}[t]
\small
\begin{center}
\begin{tabular}{l|c|c|c|c}
& dev & \multicolumn{3}{c}{test} \\ \cline{2-5}
& {\scriptsize CIDEr} & {\scriptsize CIDEr} & {\scriptsize ROUGE-L} & {\scriptsize SPICE} \\ \hline
Transf-Baseline   & - & 0.772 & 0.244 & 0.172\\
TTI-BIC (single)       & - & 0.980 & \textbf{0.266} & 0.186\\ \hline \hline
{\bf G} (Base)      & 0.868 & - & - & -\\ \hline
{\bf B}-FRCNN & 0.667 & - & - & -\\
{\bf B}-Ultra & 0.873 & - & - & -\\
{\bf L} & 0.606 & - & - & -\\ \hline
{\bf G} + {\bf B}-FRCNN          & 0.871 & - & - & - \\
{\bf G} + {\bf B}-Ultra & 0.912 & - & - & - \\ \hline
{\bf G} + {\bf L}     & 0.888 & - & - & -\\ \hline
{\bf G} + {\bf B}-FRCNN + {\bf L} & 0.892 & 0.944 & 0.261 & 0.190 \\
{\bf G} + {\bf B}-Ultra + {\bf L} & \textbf{0.937} & \textbf{0.984} & \textbf{0.265} & \textbf{0.195} \\ \hline
\end{tabular}
\vspace{-7pt}
\caption {Automatic metric scores for the image captioning task on Conceptual Captions.
Ablation results are reported for our model using different sets of visual features.
The top two baselines are from the Conceptual Captions Leaderboard as of August 30, 2019.}
\vspace{-15pt}
\label{tab:result-cap}
\end{center}
\end{table}

\paragraph{Results}

We report results on both the dev and test sets for Conceptual Captions in Table~\ref{tab:result-cap}.
``Base'' uses the {\bf G} feature only.
We first compare the Base {\bf G} against each of the feature types ({\bf B}-FRCNN, {\bf B}-Ultra, and {\bf L}).
We then perform ablations under the +{\bf B} condition (FRCNN/Ultra) to the Base {\bf G} or stronger {\bf G} + {\bf L} models.

According to dev CIDEr scores, global or box Graph-RISE features {\bf G} and {\bf B}-Ultra are (individually) clearly stronger than box features by Faster R-CNN {\bf B}-FRCNN or label embeddings {\bf L} features.
Nevertheless, these features are considerably complementary.
Specifically, box features {\bf B}-Ultra complements the Base {\bf G}, pushing the score from 0.868 to 0.912.
It is also worth noting that, albeit their low individual scores, {\bf B}-FRCNN or {\bf L} improves upon each model they are added to.

Our models with Ultra features clearly outperform the ones with FRCNN.
This is demonstrated in three conditions: when they are on their own, when they are added to the simple {\bf G} model, and when they are added to the stronger {\bf G} + {\bf L} model.
Manual inspection of the models' predictions further supports this;
a qualitative comparison of {\bf B}-Ultra vs. {\bf B}-FRCNN in Figure~\ref{fig:cap_examples} suggests that
ultra-finegrained featurization leads to an improved correspondence between visual inputs and caption tokens of \emph{unfamiliar} objects (such as ``monks" and ``staircase").

To get test scores, we submit our best model using FRCNN and our best model using Ultra (based on dev CIDEr) to the CC Challenge server.
Test scores for other models were not obtained due to the limited number of submissions per time period.
As of August 30, 2019, the {\bf G} + {\bf B}-Ultra + {\bf L} model outperforms all other single baselines\footnote{ai.google.com/research/ConceptualCaptions/leaderboard}, for both CIDEr and SPICE (and tie on ROUGE-L).

\section{Visual Question Answering}
\label{sec:vqa}

\paragraph{Dataset}
We use the recently-proposed VizWiz dataset \cite{gurari18vizwiz}, in which both images and questions originate from visually-impaired or blind people.
It consists of 20,000/3,173 $\langle$image, question, answers$\rangle$ triplets in the train/val splits, and additional 8,000 triplets for the test split.
Each question is independently annotated with 10 answers.
We choose the VizWiz benchmark specifically because it is a more suitable benchmark for measuring transfer learning effects.
Other VQA datasets, including VQA1.0~\cite{antol15vqa}, VQA2.0~\cite{goyal17vqa2}, Visual7W~\cite{zhu16visual7w}, COCOQA~\cite{ren15cocoqa}, and GQA~\cite{hudson19gqa} are completely or partly based on MSCOCO or Visual Genome.
As such, they may not provide unbiased grounds for measuring the impact of object-detection features based on Visual Genome versus alternative featurization techniques.

\paragraph{Model}
We follow the setting described in Pythia v0.1~\cite{jiang18pythia}, the winning entry to the VQA challenge 2018.
In particular, the architecture is a \emph{simplified} ``up-down" model from \cite{anderson18bottomup}\footnote{See further details in the supplementary material.}.
The featurization of the bounding boxes follows the description from Section~\ref{sec:ic}.
For the base condition, we use the box features based on Faster R-CNN ({\bf B}-FRCNN), following the majority of previous work.
For the test condition, we replace them with the Ultra-based features ({\bf B}-Ultra).

\paragraph{Metrics}
As commonly done in previous work \cite{antol15vqa}, we use as our accuracy metric the average score over 9 subsets of the ground-truth 10 answers, where each score is computed by the formula:
min(\# humans that provided that answer / 3, 1).
Accuracy on the test-dev and test-standard splits is obtained by submitting the models to the VizWiz Challenge server\footnote{{evalai.cloudcv.org/web/challenges/challenge-page/102/overview}}.

\begin{table}[!t]
\small
\begin{center}
\begin{tabular}{c|c|c|c|c|c}
 & all & y/n & num & unans & other \\ \hline
VizWiz & 46.9 & 59.6 & 21.0 & 80.5 & 27.3 \\
BAN & 51.6 & \textbf{68.1} & 17.9 & \textbf{85.3} & 31.5 \\ \hline \hline
Ours (FRCNN) & 51.9 & 66.7 & 24.3 & 85.0 & 32.1 \\
Ours (Ultra) & \textbf{53.7} & \textbf{68.1} & \textbf{28.8} & 84.0 & \textbf{35.4} \\ \hline
\end{tabular}
\vspace{-7pt}
\caption {Accuracy (\%) on the test-standard split for the VQA task on the VizWiz dataset.
Additionally, we provide accuracy per answer type: yes/no (y/n), number (num), unanswerable (unans), and the rest (other).
The baselines include VizWiz \cite{gurari18vizwiz} and BAN \cite{kim18banvizwiz}.}
\vspace{-10pt}
\label{tab:results-vqa}
\end{center}
\end{table}

\paragraph{Results}
We report results on the VizWiz benchmark in Table~\ref{tab:results-vqa}.
Our model with FRCNN provides a strong baseline, slightly outperforming the previous-best model, BAN~\cite{kim18banvizwiz}, a different architecture that also uses the FRCNN-based features for object bounding boxes.
The model using Ultra features further improves upon this; at 53.7\%, it outperforms the one using FRCNN by a significant margin (1.8\% accuracy on ``all'' question types).
Moreover, this 1.8\% improvement is a weighted average across answer types; the per-answer-type numbers indicate that our approach achieves even better improvements on two of the more difficult answer types, ``number" (+4.5\%) and ``rest" (+3.3\%).
These improvements are illustrated by the examples provided in Figure~\ref{fig:examples}.

This illustrates the effectiveness of decoupling bounding box proposal and featurization, and quantifies the impact of using transfer learning via large amounts of training data and ultrafine-grained semantic labels used for object representations.

\section{Conclusion}

In this work, we propose to (re)decouple box proposal and featurization.
We show that this allows us to leverage additional signals and annotations,
leading to more effective transfer learning for downstream vision and language tasks: image captioning and visual question answering.
This result suggests that large-scale datasets with fine-grained image-level semantic labels,
even when they do not dissect complex visual scenes, can benefit current state-of-the-art models -- especially when applied to benchmarks where images are from diverse domains.

\bibliography{main}
\bibliographystyle{acl_natbib}

\newpage
\appendix
\section{Details on Faster R-CNN}
Our model is implemented in TensorFlow \cite{tensorflow2015-whitepaper}.
We follow \newcite{anderson18bottomup} in terms of model architecture, data splits, and processing steps.
We describe major components and differences below.
In particular, we use the latest version of Visual Genome (v1.4), with 1600 object and 400 attribute categories.
We also have the ``background" class for objects and the ``no attribute" class for attributes.
We limit the number of attributes per object to 16.
We resize the image to so that the maximum of height or width is 896.
We train our model with a batch size of 64 for 50K steps, using SGD with momentum on an 8-core Google Cloud TPU\footnote{cloud.google.com/tpu}.
We clip the gradient if the norm is greater than 10.
We use the cosine learning rate schedule with 1K warm-up steps, increasing the learning rate from 0.003 to 0.04 and reducing it to 0.01 at step 20K and to 0.005 at step 40K.
We apply random crops to images and use batch normalization \cite{ioffe15batch} as well as DropBlock \cite{ghiasi18dropblock} on block 3 and block 4 of the ResNet-101 during training.
Our features come from fc6 after ReLU.

\section{Details on Image Captioning}

\begin{figure*}[ht]
\begin{center}
 \includegraphics[width=1.0\linewidth]{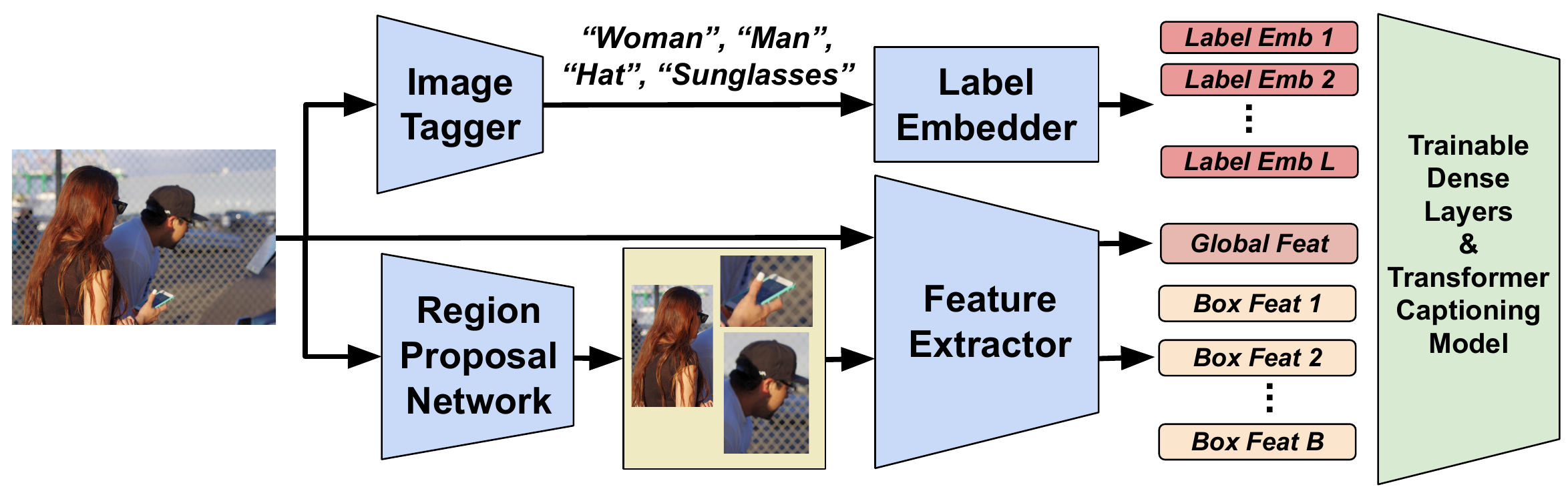}
\end{center}
 \caption{Pipeline for converting an image to a sequence of image features in our highest performing image captioning model on the Conceptual Captions benchmark, used as input to the Transformer-based model.}
\label{fig:image_features}
\end{figure*}

Our model is implemented in TensorFlow \cite{tensorflow2015-whitepaper}.
Our Transformer-based architecture has a stack of 6 layers for both the encoder and the decoder.
The number of attention heads is set to 8.
We do not use positional encoding.
We have an additional dense projection layer for each type of input features (see Figure~\ref{fig:image_features} for examples).
Moreover, for Faster-RCNN features, we observe the best performance when first transforming the 2048D input feature vector to a 64D one (as in Ultra) using another projection layer, and thus report accuracy numbers in this setting.
At the same time, we also have these projection layers in our VQA architecture when using Ultra features (see the next section).
We use Adam optimizer \cite{adam} with a warm-up style learning rate schedule, linearly increasing the learning rate in the first 20 epochs until it reaches 0.000032 and then use a decay rate of 0.95 for every 25 epochs.
We tuned the initial learning rate over \{0.000016, 0.000032, 0.000064\}.
We train our model with a batch size of 4096 on a 32-core Google Cloud TPU for a total of 2 million steps.
Each training run takes approximately 4 days.

In Figure~\ref{fig:image_features}, we show how we convert an image (pixels) to an input sequence of image features to the Transformer-based model described in the main text.

\section{Details on VQA}

\begin{table*}[ht]
\small
\begin{center}
\begin{tabular}{c|c|c|c|c|c|c|c|c|c|c|c}
 & val & \multicolumn{5}{c|}{test-dev} & \multicolumn{5}{c}{test-standard} \\ \hline
 & all & all & y/n & num & unans & other & all & y/n & num & unans & other \\ \hline
VizWiz \cite{gurari18vizwiz} & - & - & - & - & - & - & 46.9 & 59.6 & 21.0 & 80.5 & 27.3 \\
BAN \cite{kim18banvizwiz} & - & - & - & - & - & - & 51.6 & \textbf{68.1} & 17.9 & \textbf{85.3} & 31.5 \\ \hline \hline
Ours (FRCNN) & 55.2 & 53.6 & 72.7 & 22.7 & 85.9 & 33.3 & 51.9 & 66.7 & 24.3 & 85.0 & 32.1 \\
Ours (Ultra) & 56.8 & 55.1 & 71.7 & 31.6 & 84.4 & 36.7 & \textbf{53.7} & \textbf{68.1} & \textbf{28.8} & 84.0 & \textbf{35.4} \\ \hline
\end{tabular}
\vspace{-7pt}
\caption {Accuracy (\%) for the VQA task on the VizWiz dataset.
Additionally, we provide accuracy per answer type on the test-dev and test-standard splits: yes/no (y/n), number (num), unanswerable (unans), and the rest (other).}
\vspace{-10pt}
\label{tab:supp-results-vqa}
\end{center}
\end{table*}

Our model is implemented in TensorFlow \cite{tensorflow2015-whitepaper}.
As mentioned in the main text, the architecture is a \emph{simplified} ``up-down" model of \cite{anderson18bottomup}.
This architecture has two major differences.
First, it uses weight normalization \cite{salimans16weightnorm} followed by ReLU instead of the more expensive gated hyperbolic tangent activation.
Second, it uses multi-modal combination by element-wise multiplication instead of by feature concatenation.

The only minor differences from Pythia v0.1 are that we use Adam \cite{adam}, not its variant AdaMax,
and that we use a single classifier layer instead of two vision and language layers in all of our experiments.
We use Pythia v0.1 \cite{jiang18pythia} to preprocess VizWiz dataset and retain 3135 top answers.
Analogous to what we observe in our image captioning model, for Ultra features, we see the best performance when scaling and expanding the 64D input feature vector to a 2048D one (as in FRCNN) using another projection layer, followed by ReLU.
We thus report accuracy numbers in this setting.
We use a warm-up style learning rate schedule, linearly increasing the learning rate in the first 10 epochs until it reaches the initial learning rate, and then use a decay rate of 0.5 for every 20 epochs.
We tune the initial learning rate over \{0.00005, 0.0001, 0.0003, 0.0005, 0.001, 0.003\}.
We train our model with a batch size of 192 on an 8-core Google Cloud TPU for a total of 70K steps.
Each training run takes approximately 2 hours.

\section{Full results on the VizWiz benchmark}

Table~\ref{tab:supp-results-vqa} reports accuracy on additional splits of VizWiz, complementing the one in the main text.

\end{document}